\documentclass[runningheads]{llncs}

\usepackage{eccv}

\usepackage{eccvabbrv}

\usepackage{graphicx}
\usepackage{booktabs}

\usepackage[accsupp]{axessibility}  % Improves PDF readability for those with disabilities.

\usepackage{hyperref}

\usepackage{orcidlink}
\usepackage{marvosym}
\hypersetup{colorlinks = true,
	    urlcolor = blue
           } 
\definecolor{mycolor_grey}{RGB}{208,206,206} 
\definecolor{mycolor_green}{RGB}{227,242,217} 
\usepackage{wrapfig}
\usepackage{bm}
\usepackage{threeparttable}
\usepackage{multirow}
\usepackage{floatflt}
\usepackage{graphicx}

\begin{document}

% ---------------------------------------------------------------
% TODO REVIEW: Replace with your title
\title{Training A Small Emotional Vision Language Model for Visual Art Comprehension} 

% TODO REVIEW: If the paper title is too long for the running head, you can set
% an abbreviated paper title here. If not, comment out.
\titlerunning{Training A SELVM for Visual Art Comprehension}

% TODO FINAL: Replace with your author list. 
% Include the authors' OCRID for the camera-ready version, if at all possible.
\author{Jing Zhang\inst{1}\orcidlink{0009-0005-1590-5886} \and
Liang Zheng\inst{3}\textsuperscript{(\Letter)}\orcidlink{0000-0002-1464-9500} \and
Meng Wang\inst{1,2}\orcidlink{0000-0002-3094-7735} \and
Dan Guo\inst{1,2}\textsuperscript{(\Letter)}\orcidlink{0000-0003-2594-254X}}

% TODO FINAL: Replace with an abbreviated list of authors.
\authorrunning{J.~Zhang et al.}
% First names are abbreviated in the running head.
% If there are more than two authors, 'et al.' is used.

% TODO FINAL: Replace with your institution list.
\institute{Hefei University of Technology, Hefei, China \and Institute of Artifcial Intelligence, Hefei Comprehensive National Science Center \\
\email{hfutzhangjing@gmail.com, eric.mengwang@gmail.com, guodan@hfut.edu.cn}\\
\and
Australian National University, Canberra, Australia \\
\email{liang.zheng@anu.edu.au}} 

\maketitle

\setcounter{footnote}{0}

\begin{abstract}
    This paper develops small vision language models to understand visual art, which, given an art work, aims to identify its emotion category and explain this prediction with natural language. While small models are computationally efficient, their capacity is much limited compared with large models. To break this trade-off, this paper builds a small emotional vision language model (SEVLM) by emotion modeling and input-output feature alignment. On the one hand, based on valence-arousal-dominance (VAD) knowledge annotated by psychology experts, we introduce and fuse emotional features derived through VAD dictionary and a VAD head to align VAD vectors of predicted emotion explanation and the ground truth. This allows the vision language model to better understand and generate emotional texts, compared with using traditional text embeddings alone. On the other hand, we design a contrastive head to pull close embeddings of the image, its emotion class, and explanation, which aligns model outputs and inputs. On two public affective explanation datasets, we show that the proposed techniques consistently improve the visual art understanding performance of baseline SEVLMs. Importantly, the proposed model can be trained and evaluated on a single RTX 2080 Ti while exhibiting very strong performance: it not only outperforms the state-of-the-art small models but is also competitive compared with LLaVA 7B after fine-tuning and GPT4(V). The code is available at \href{https://github.com/BetterZH/SEVLM-code}{https://github.com/BetterZH/SEVLM-code}.
\end{abstract}

\section{Introduction}
\label{sec: Introduction}
\begin{figure}[t]
	\centering
 \includegraphics[width=1.0\columnwidth]{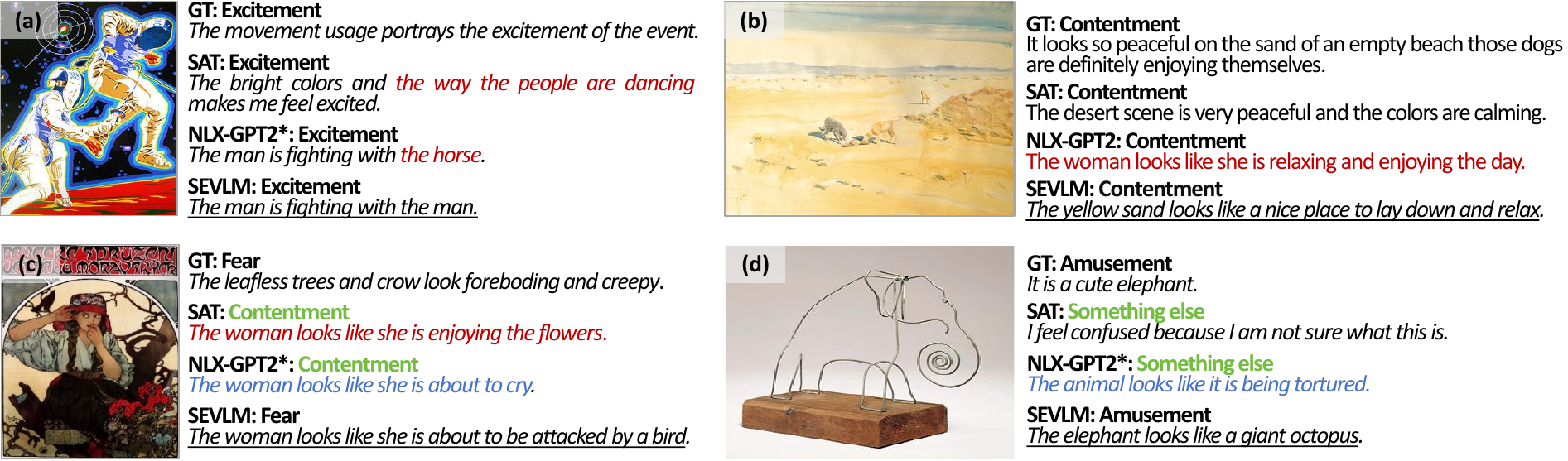}
  \caption{Examples comparing different methods of predicting emotion class and explaining why this emotion is evoked given an art image on both ArtEmis v1.0 test set  and ArtEmis v2.0 Combined test set. Three models are compared: SAT \cite{achlioptas2021artemis}, NLX-GPT2 \cite{sammani2022nlx}, and our method. In both examples, the explanations from existing methods are misaligned with the emotion label or the art image, but our method gives superior results. \textcolor[RGB]{117,189,66}{\textbf{Green fonts}} indicate incorrect emotion classification results; \textcolor[RGB]{192,0,0}{\textbf{red texts}} indicate large discrepancies between the semantics of explanations and visual content; \textcolor[RGB]{72,116,203}{\textbf{blue texts}} denote that the emotion of the explanations does not correspond to the predicted category. Our design in \cref{sec: Proposed Improvements} aims to alleviate these problems.}
	\label{fig1:comparison}
\end{figure}

Understanding the emotion of a human viewing visual art works can be a milestone for vision language models. This problem is challenging because art is abstract and it is subjective to explain feelings. This paper studies a specific emotion understanding problem: given an art image, the system identifies the emotion category \eg, `contentment', and provides a language explanation, \eg, `the yellow sand looks like a nice place to lay down and relax' in \cref{fig1:comparison} (b).

An appealing way of addressing this problem is to fine-tune a large vision language model such as LLaVA \cite{liu2023llava}, using training art images and their manual annotations. This method, as to be shown in our experiment (see \cref{sec: main evaluation}), is useful. But its down side is also obvious: with billions of model parameters, the computational cost is high. In comparison, Small Vision Language Models (SVLMs), with one-two orders of magnitude less parameters, present a much more efficient solution, but they have limited model capacity. 

\begin{figure}[t]
	\centering
 \includegraphics[width=1\columnwidth]{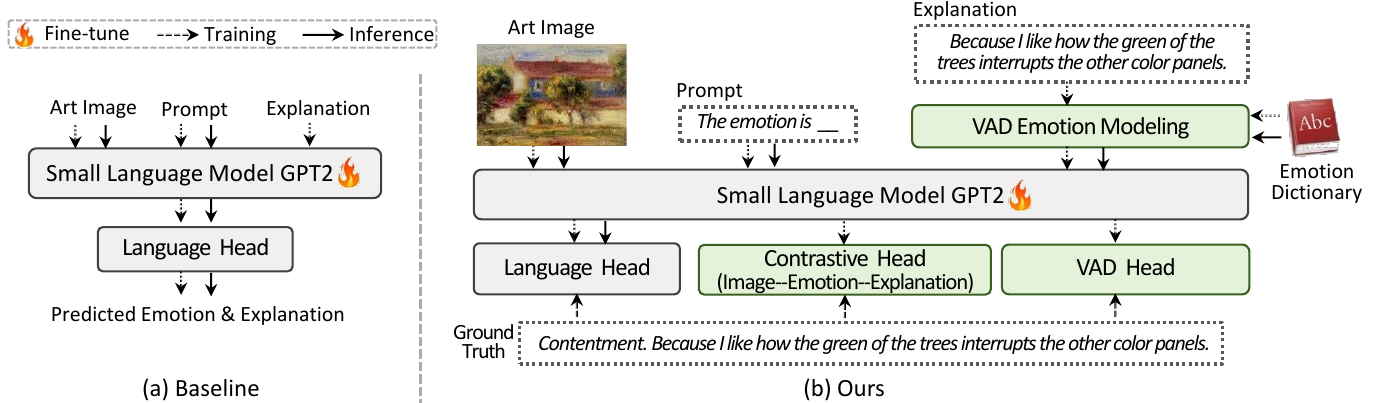}
  \caption{Overview and comparison between the baseline model (a) and the proposed model (b). The \colorbox{mycolor_grey!30}{gray} boxes in (a) and (b) are the same: an art image, a prompt, and an explanation are used as input to GPT2 and then a language head, which will output an emotion class and corresponding explanations. The \colorbox{mycolor_green}{green} boxes in (b) denote our technical contributions. We use a VAD dictionary to provide emotion features which is complementary to the standard text embeddings. Moreover, we design a VAD head and a contrastive head to facilitate emotion learning and feature alignment among the image, emotion class and explanation, respectively.}.
	\label{fig2:baseline}
\end{figure}

We are interested in building a small emotional vision language model (SEVLM) to break the second trade-off, \ie, improving the art understanding ability of small vision language models while maintaining their computational efficiency. A baseline approach would be fine-tuning a small language model, \eg, GPT2 \cite{sammani2022nlx,mokady2021clipcap} with affective explanation training data, as shown in~\cref{fig2:baseline} (a). However, we find this baseline makes two major mistakes in practice. On the one hand, because the language models are pre-trained on objective and precise text descriptions, the language explanations of the viewer emotion are often not emotional and subjective (red texts in \cref{fig1:comparison}). On the other hand, the image, emotion class, and the explanations are often misaligned with each other (blue and green texts in \cref{fig1:comparison}).

To make the output emotion explanations more emotional, we borrow the idea of valence-arousal-dominance (VAD) modeling: each word in the explanation text is represented by a 3-dim vector defined in the VAD dictionary \cite{mohammad2018obtaining}. Because the VAD scheme provides rich psychological descriptions of the text, we 1) fuse the VAD text features with the classic text features to improve the model input\footnote{In image captioning \cite{achlioptas2021artemis,mohamed2022okay,xu2015show}, language explanations are used both as input and output during training but only as output in inference.}, and 2) design a VAD head to enforce the output text explanation to have similar VAD vectors with those of the ground truth explanation. The two measures allow the language model to better understand and output emotional texts.

To improve the alignment between the image, emotion category, and explanations, we design a constrastive head to enforce features of the image, emotion label, and text explanation have similar embeddings. This is implemented by a standard contrastive learning loss. Our experiments show that the above techniques consistently improve emotion understanding capacity of the vision language model. Main points of this paper are summarized below. 
\begin{itemize}
\item 
We present a small emotional vision language model (SEVLM) to understand emotion in visual art. Its accuracy is superior to state-of-the-art small models and GPT4(V), and on par with fine-tuned large models. Computationally, our model can be trained and tested on a single RTX 2080 Ti GPU. 
\item Technical contribution 1: We borrow the VAD dictionary from psychology to provide emotion-aware text features besides the classical text embedding. 
\item Technical contribution 2: We propose a VAD head to align the VAD vector of the system output with that of the ground truth. Together with the first contribution, it enables emotional outputs. 
\item Technical contribution 3: We propose a contrastive head to align features of the image, emotion class, and explanation features.
\end{itemize}

\section{Related Work} 
\label{sec: related work}

\textbf{Visual emotion understanding} has been long studied, where emotion classification is particularly well-known \cite{you2015robust,yang2021stimuli,chen2014deepsentibank,cen2024masanet,xu2022mdan}. Recently, {emotional image captioning (EIC)} gained increasing attention. The EIC models \cite{mathews2016senticap,zhao2020memcap,li2021image,wu2023sentimental} focus on describing the visual content with affective words (\eg, `lovely' or `alone'), aiming to enhance the attractiveness and distinctiveness of text descriptions. Differently, we focus on interpreting emotion class prediction from images \cite{achlioptas2021artemis,mohamed2022okay}.

\textbf{Emotional language models.} Large language models (LLMs) have impressive capabilities in generic fields, such as coding and chatting. But they may be limited in the verticle domain of emotion understanding. DialogueLLM \cite{zhang2023dialoguellm} is an early work among the few in this area, which is designed for emotion recognition in conversations by fine-tuning LLMs with multimodal (\ie, texts and videos) emotional dialogues. {Dfferently, our work gives attention to emotional art understanding by reasoning the cause behind emotion choice, which offers an orthogonal view in emotional language models.}

\textbf{\textbf{AI in art understanding}} includes cross-modal retrieval \cite{ananthram2023feelingblue}, visual question answering \cite{garcia2020dataset} and image captioning \cite{bai2021explain,lu2022data,ruta2022stylebabel}. Recent works \cite{achlioptas2021artemis,mohamed2022okay,achlioptas2023affection} study the emotion response of viewers and why such emotion is evoked from an artwork. Achlioptas \etal \cite{achlioptas2021artemis} introduce an affective explanation dataset `ArtEmis'. They also develop a two-stage method where classification and explanation networks are \textit{small} and \textit{separate}. Specifically, it first predicts emotion category by an emotion recognition model \cite{he2016deep} and then uses this prediction together with the art image as the input of a caption model \cite{xu2015show} to produce emotion explanation. A subsequent work \cite{mohamed2022okay} uses data augmentation to enhance the image captioning model so also has two stages. In comparison, the \textit{small} model we develop is \textit{end-to-end} and has \textit{superior performance}.

\section{Preliminaries: Word to VAD Vector}
\label{sec: preliminaries}
\begin{wrapfigure}[14]{r}{0.29\textwidth}
	\centering
 \includegraphics[width=0.29\textwidth]{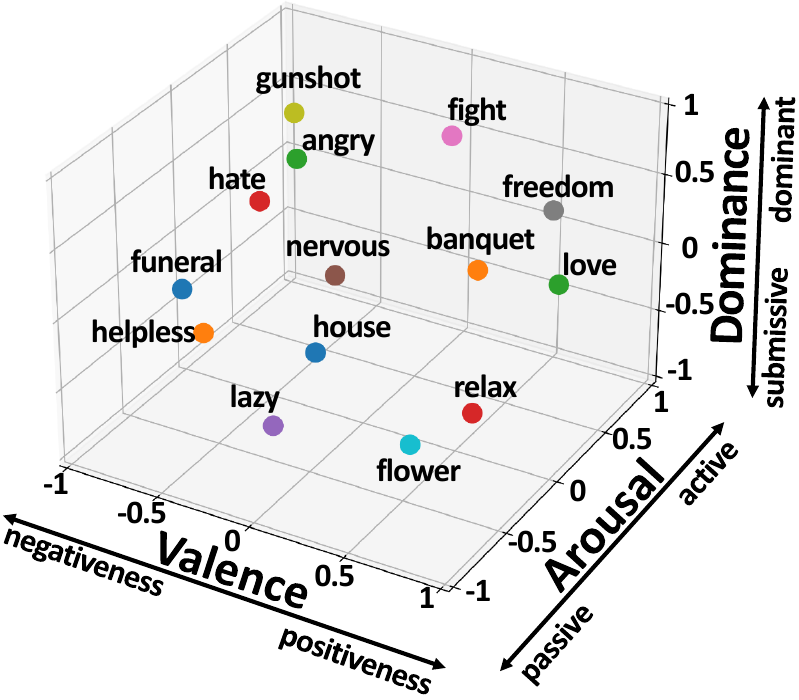}
  \caption{Depicting VAD vectors of example words in the 3-dim space.} \label{fig3:vad}
\end{wrapfigure}
The VAD lexicon \cite{mohammad2018obtaining} from National Research Council Canada is a public language tool to obtain the VAD word vectors. This dictionary presents human ratings of valence (positiveness–negativeness), arousal (active–passive), and dominance (dominant–submissive) for more than 20,000 English words \cite{mohammad2018obtaining,russell2003core}. In dictionary, the 3-dim vector $(v,a,d)^T \in \mathbb{R}^3$ for every word contains real-valued numbers in the interval between -1 (lowest V, A, or D) and 1 (highest V, A, or D). For example, as shown in \cref{fig3:vad}, the VAD vector of the word \emph{banquet} is $(0.53, 0.142, 0.2)^T$, while that of \emph{funeral} is $(-0.854, -0.24, -0.214)^T$. It means that \emph{banquet} represents greater positivity, higher arousal, and more dominance than \emph{funeral}. For words outside the dictionary, we consider them as neutral emotion words, and set $(v,a,d)=(0,0,0)^T$. This work shows that the VAD dimensions of word are beneficial for emotion analysis.

\section{Approach}
This section describes the Small Emotional Vision Language Model (SEVLM) for visual art appreciation. As shown in \cref{fig2:baseline} and \cref{fig4:detailed network}, it mainly consists of four parts. 1) A {vision language backbone} in the baseline that consists of an image encoder, a GPT2 decoder and a traditional language head, to be described in \cref{sec:Baseline}. 2) The {VAD emotion modeling} introduces emotion knowledge using \emph{V}alence-\emph{A}rousal-\emph{D}ominance into input text embedding to enhance the emotion understanding ability of our model. 3) A {VAD head} is devised for VAD-aware emotion explanation generation. 4) A {contrastive head} is used to align features among the image, emotion label, and explanation text. The latter three techniques will be described in~\cref{sec: ternary contrastive loss}.

\subsection{Baseline Structure: A Small Language Model} % and Inference}
\label{sec:Baseline}

\textbf{Overview.} 
As shown in \cref{fig2:baseline} (a), our baseline has three inputs: An art image, an emotion classification prompt $M$: 
`\emph{the emotion is \_}', and the ground truth test explanation text $X$. %The test explanation is used as both input and output in training and output only in inference. 
Outputs of the baseline are the predicted emotion category and language explanation $X$ of the prediction. Our system is composed of an image encoder, a text encoder, and a language decoder. 

\textbf{Basic components.} We use CLIP vision encoder \cite{radford2021learning} as the \underline{image encoder} with frozen parameters. It encodes image $I$ and outputs feature $\bm{f}^I$ $\in \mathbb{R}^{K*d_v}$, where $K$ is the number of image patches, and $d_v$ is the patch feature dimension.

\begin{figure*}[t]
	\centering
	\includegraphics[width=1.0\linewidth]{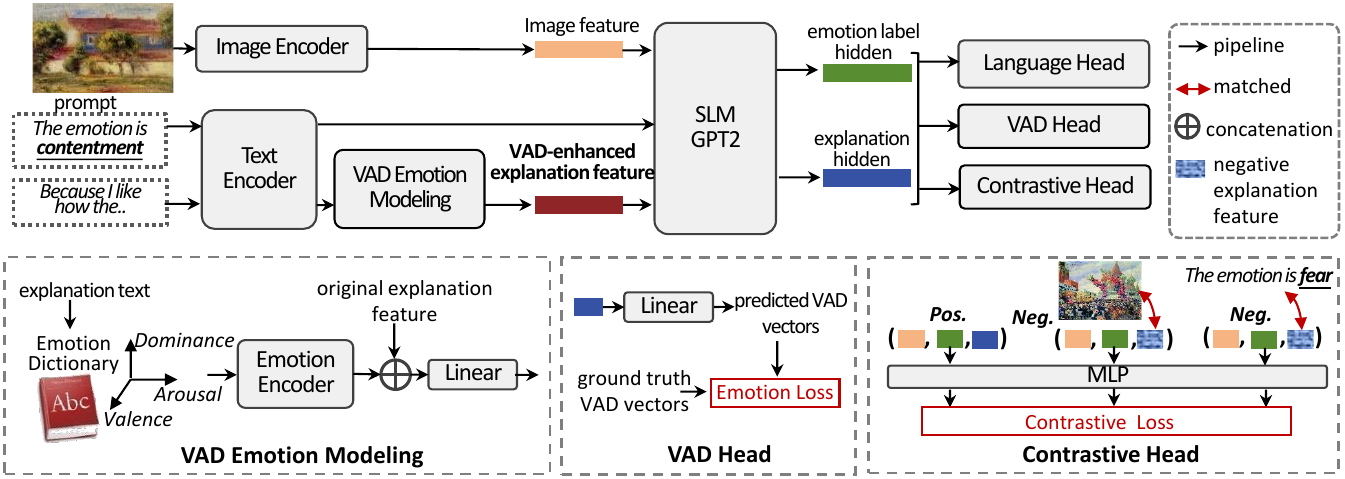}
    \caption{Detailed network structure of the proposed small emotional vision language model. It has: 1) a vision language backbone including an image encoder, a samll language model (SLM) GPT2 decoder, and a language head; 2) VAD emotion modeling introducing emotion knowledge VAD into text embeddings to enhance model capacity of understanding emotion; 3) a VAD head  to learn VAD-aware emotion; and 4) a contrastive head to force the features alignment among image, emotion label and explanation. During training, we use the emotion label and explanation as ground truth. In inference, we use the prompt `The emotion is \_' and an art image as input and generate the emotion label and explanations. 
 }
\label{fig4:detailed network}
\end{figure*}

The \underline{text encoder} consists of a word embedding layer, a position embedding layer \cite{vaswani2017attention}, and a segment embedding layer\cite{sammani2022nlx} (see \cref{fig4:detailed network}). The word embedding layer converts each token of input text into a vector of $d_s$ dimension, where $d_s$ denote the dimension of word embedding. The position embedding layer is used to encode positional information of input sentence. {The segment embedding layer encodes two types of tokens,} \ie, [$M$: $\left \langle emotion \right \rangle$, $X$: $\left \langle explanation\right \rangle$]. During training, we use the concatenation of emotion classification prompt $M$ and text explanation $X$, named \emph{full sentence}, as input of text encoder. The dimensions of the three vectors output by these three embedding layers are all $L*d_s$, where $L$ is the length of input text. By summing the outputs of the above, we obtain the feature representation $\bm{f}^{S}=\bm{f}^{M} \oplus \bm{f}^{X} \in \mathbb{R}^{L*d_s}$, where $\bm{f}^{M}$ and $\bm{f}^{X}$ are textual embeddings of emotion sentence $M$ and explanation sentence $X$, respectively, and $\oplus$ denotes matrix concatenation, and $L$ is the length of \emph{full sentence}.

GPT2 \cite{radford2019language} is chosen as the \underline{language decoder}. It takes as input features of the art image and full sentence, \ie, $\bm{f}^I$ and $\bm{f}^{S}$, respectively, and predicts the emotion class and explanation.

\textbf{Adding cross-attention to fuse image and text.} GPT2, mainly consisting of self-attention layers, is not originally designed for multi-modal inputs. To improve this, we introduce cross attention in each block. Textual feature $\bm{f}^{S}$ and the visual embedding $\bm{f}^{I}$ are fed into the GPT2 decoder, yielding the hidden states $\bm{f'}^{M}$ and $\bm{f'}^{X}$ . The process is formulated as:
\begin{equation}
\bm{f'}^{M},\bm{f'}^{X} ={\rm GPT2Decoder}(\bm{f}^S,\bm{f}^I),
\end{equation}
where $\bm{f'}^{M}$ and $\bm{f'}^{X}$ correspond to % the hidden states of 
emotion sentence and explanation sentence, respectively. We denote full sentence hidden states $\bm{f'}^{S} =\bm{f'}^{M} \oplus \bm{f'}^{X} \in \mathbb{R}^{L*d_s}$.

\textbf{Loss.} 
A standard \underline{language head} maps the full sentence hidden stat $\bm{f'}^{S}$ to the vocabulary space. We use the cross-entropy objective as the \emph{language loss} to generate emotion class and text explanation. 

During inference, we only use the image and prompt $M$ as input to generate emotion label and explanation (refer \cref{fig2:baseline}).

\subsection{Proposed Improvements}
\label{sec: Proposed Improvements}
\subsubsection{VAD emotion modeling.}\label{sec: VAD Embeddings}
The emotional label in emotion sentence $M$ possesses a distinct emotional coloring, \eg, `fear' and `awe', while the words in the explanation $X$ may be general and not as emotionally colored, \eg, `how' and `interrupt'. To make the text explanation suitably emotional, we use VAD vectors $\{(v_t,a_t,d_t)\}_{t \in T}$ for the explanation text to supplement the standard text embedding, where $T$ is the length of explanation. 

For VAD emotion modeling, we design an \underline{emotion encoder} using a transformer encoder \cite{vaswani2017attention}. It encodes VAD vectors $\{(v_t,a_t,d_t)\}_{t \in T}$ into a 3-dim emotion feature $\bm{f}^{E} \in \mathbb{R}^{T*3}$. Then, we concatenate it with the classic text embedding $\bm{f}^{X} \in \mathbb{R}^{T*d_s}$ along the feature dimension to obtain a combined feature of $T*(d_{s}+3)$ dimensions. A single linear layer maps the combined feature to the input embedding space of GPT2 decoder. The whole process is formulated as: 
\begin{align}\left\{\begin{aligned}\label{eq: emotion-enhanced f_X}
	&\bm{f}^{E}\!=\!{\rm EmotionEncoder} (\{(v_t,a_t,d_t)\}_{t \in T}) \\
       &\bm{\hat{f}}^{X} = W^{E}(\bm{f}^{X} \oplus \bm{f}^{E})+b^{E}
\end{aligned},\right.\end{align}
where $\bm{\hat{f}}^{X} \in \mathbb{R}^{T*d_s}$ is the \underline{VAD-enhanced explanation feature}, $W^{E}$ and $b^{E}$ are learnable parameters, and $\oplus$ denotes matrix concatenation. 

The updated embedding of \emph{full sentence} is formulated as $\bm{\hat{f}}^{S} = \bm{f}^{M} \oplus \bm{\hat{f}}^{X} \in \mathbb{R}^{L*d_s}$. Outputs of the GPT2 decoder are:
\begin{equation}\label{eq:}
\bm{f'}^{M},\bm{f'}^{X} ={\rm GPT2Decoder}(\bm{\hat{f}}^S,\bm{f}^I),
\end{equation}
where hidden states of the full sentence are $\bm{f'}^{S} =\bm{f'}^{M} \oplus \bm{f'}^{X} \in \mathbb{R}^{L*d_s}$.

\subsubsection{VAD head.}
\label{sec: emotion loss}
The language head in the baseline (refer \cref{sec:Baseline}) enforces explanation generation from a generic point of view. To further ensure the explanation incorporates VAD knowledge from the VAD emotion modeling method, we propose a \underline{VAD head},
which is implemented as a single linear layer. At each decoding time $t$, the predicted VAD vector of each word in explanation is determined by:
\begin{equation}
\bm{f'}^{E}_t= W^{V}\bm{f'}^{X}_{t} + b^{V},
\end{equation}
where ${\bm{f'}}^{E}_t=(v'_t,a'_t,d'_t)$, $\bm{f'}^{X}_t$ is the $t$-th hidden state of $\bm{f'}^{X}$, and $W^{V}$ and $b^{V}$ are learnable weights and biases. 

We use mean squared error (MSE) as the \emph{emotion loss}, minimizing the differences between predicted VAD vectors and ground truth VAD vectors:  
\begin{equation}
\mathcal{L}_{emotion} = \frac{1}{T} \sum_{t=1}^{T} (\bm{f}^{E}_t -{\bm{f'}}^{E}_t)^2,
\end{equation}
where $\bm{f'}^{E} =\{(v'_t,a'_t,d'_t)\}_{t \in T} \in \mathbb{R}^{T*3}$ is the generated VAD vectors of explanation $X{'}$, and $T$ is the length of $X{'}$.

\subsubsection{Contrastive head.}
\label{sec: ternary contrastive loss}
We observe that the explanations are often misaligned with the emotion label and the art image (see \cref{fig1:comparison}). To solve this problem, we propose a \underline{contrastive head} to align features of the three entities.

It is implemented as a multilayer perceptron (MLP), taking visual, emotional, and explanatory features as inputs and outputs a score. This score is used for evaluating the alignment among these three aspects, with higher scores indicating a better match. We define this similarity score as $\mathcal{S}$, formulated as:
\begin{equation}
\mathcal{S}(I, M, X) = {\rm MLP}( \mu( \bm{f}^I)\oplus \mu (\bm{f'}^{M}) \oplus \mu (\bm{f'}^{X})),
\end{equation}
where $\mu (\cdot)$ denotes the operation of taking the mean, outputting a vector of $1*d_s$, and $\oplus$ denotes concatenation.

We use a standard contrastive learning objective as our \emph{contrastive loss}. We train the model under the objective that the score of matched ternary features should be higher than that of unmatched ternary features: 
\begin{equation}
\mathcal{L}_{contrastive} \!\!= \!\!-\!\! \sum_{b \in B}\! \frac{ e^{\mathcal{S}_{b}(\!I\!, M\!, X\!) } } {  e^{\mathcal{S}_{b}(\!I\!, M \!, X\!)} + e^{\mathcal{S}_{b}(\!I\!, M\!, X_{\neq \! M}\!)}  + e^{\mathcal{S}_{b}(\!I\!, M\!, X_{\! \neq I}\!)}  },
\end{equation}
where $B$ denotes the batch size, and $X_{\!\neq \!M}$ denotes a negative sample with a wrong emotion label, while $X_{\!\neq I\!}$ denotes a negative sample from other images. 

Our model is trained by minimizing the weighted sum of all losses: 
\begin{equation}\label{eq: all losses}
    \mathcal{L} = \mathcal{L}_{language} + \mathcal{L}_{emotion} + \alpha \mathcal{L}_{contrastive}.
\end{equation}
Here $\alpha$ is hyper-parameter, whose impact is evaluated in the supplementary material. % \cref{sec: Further Analysis}.

\subsection{Novelty and Contribution Statement}
The VAD dictionary was released to provide words with expert emotion annotations. It has been used in traditional sentiment analysis problems such as recognising emotion classes from texts \cite{yang2023cluster,zhong2019knowledge} and detecting humor moments (binary classification) from multi-modal inputs \cite{hasan2021humor}. Yet, it largely remains unknown how it benefits generative models especially in the emotion explanation domain. This paper bridges this gap using VAD modeling to improve both the input text embeddings and loss function of vision language models. This may spark further exploration of these expert annotations. 

On the other hand, the contrastive loss is traditionally applied for a pair of data, such as a pair of images \cite{chen20simple,jung2022exploring} or image and text \cite{liu2022multimedia,radford2021learning}. The triplet loss, while using sample triplets, usually deals with heterogeneous ones, \eg, image triplets \cite{wang23semantic,chen2017beyond}.
Differently, in the field of emotion explanation, three heterogeneous features should be aligned, \ie, emotion category, text explanation, and the input image. The proposed (ternary) contrastive loss thus provides a unique mechanism to compute the alignment score among three sample types to improve input-output alignment. This insight may be very useful for other tasks with multiple and correlated inputs and outputs.

\section{Experiments}

\subsection{Experimental Setup}
\textbf{Datasets.} We conduct experiments on two benchmark datasets: ArtEmis v1.0 \cite{achlioptas2021artemis} and ArtEmis v2.0 \cite{mohamed2022okay}. ArtEmis v1.0 dataset comprises 80,031 fine art paintings and ArtEmis v1.0/v2.0 includes 454,684/455,000 affective responses with explanatory utterances. Following existing works \cite{achlioptas2021artemis,mohamed2022okay}, we use an \{85$\%$, 5$\%, $10$\%$\} split for $\{$training, validation, testing$\}$. And we follow the emotion category set from Ekman emotion categories \cite{ekman1992argument}, \ie, the emotion label in $M \in \{$`amusement', `awe', `contentment', `excitement', `fear', `sadness', `anger', `disgust'$\}$. 

\textbf{Evaluation metrics.} For emotion classification, we use accuracy (\textbf{ACC}) as the evaluation metric. ACC refers to the ratio of predicted emotion that aligns with the dominant emotion of the image \cite{achlioptas2021artemis}. For explanation, we utilize a few popular machine-based metrics, including \textbf{BLEU}, \textbf{METEOR}, and \textbf{ROUGE} (abbreviated as B, M, R), to evaluate the semantic relevance of generated explanations. We also use emotion-alignment (\textbf{EA}) \cite{achlioptas2021artemis} to measure whether the deduced emotion of explanation is aligned with the dominant emotion of the image. The \textbf{Unique} metric is used to assess the proportion of distinct generated explanations in the test set, indicating the explanation diversity.

\textbf{Implementation Details.} We use CLIP ViT-B/16 \cite{radford2021learning} as image encoder. The visual feature dimensions are set to $K=196$ and $d_v=768$. We choose GPT2 \cite{sanh2019distilbert} as language model. It consists of $N=6$ transformer blocks and 12 attention heads. The emotion encoder is configured with three transformer blocks and one attention head. Across all modules in our model, the word embedding dimension $d_s$ is consistently set to 768. During training, each \emph{full sentence} starts with $\left \langle bos\right \rangle$ and ends with $\left \langle eos\right \rangle$. The length $L$ of is set to 30.

During training, we adopt the AdamW optimizer \cite{gugger2018adamw} with a learning rate of 2e-5 for both GPT2 decoder and three heads, and a learning rate of 4e-5 for VAD emotion modeling. Batch size $B$ is set to 32. The hyperparameter $\alpha$ is set to 2 for both datasets v1 and v2. During testing, we use nucleus sampling \cite{holtzman2019curious} with a probability $0.9$. For detailed computational costs, please refer to \cref{tab: Computational Cost}.

\subsection{Main Evaluation} \label{sec: main evaluation}
\begin{table*}[t]
	\caption{Method comparison on the ArtEmis v1.0 test set and ArtEmis v2.0 Combined test set. Our model is superior in both emotion classification and explanation tasks. * indicates results reproduced by us. The best and second best numbers in each column are marked with bold font and underlined, respectively.}
    \scriptsize
	\centering
        \begin{threeparttable}
    \begin{tabular}{c|l|c|c|c|ccccccc}
                    \toprule
                   Dataset & Method & Backbone & ACC$\uparrow$ & EA$\uparrow$ &  B1$\uparrow$ &   B2$\uparrow$  & B3$\uparrow$ & B4$\uparrow$ & M$\uparrow$ & R$\uparrow$ & Unique$\uparrow$ \\
                    \midrule
                \multirow{5}{*}{v1.0} & M2 \cite{achlioptas2021artemis} & Trans-Trans & \multirow{3}{*}{60.2} & 52.1 & 51.1 & 28.2 & 15.4 & \underline{9.0} & \underline{13.7} & 28.6   & 23.0 
                \\
                 ~ & SAT \cite{achlioptas2021artemis}
                  & CNN-LSTM &  ~ & 51.9 & 52.0  & 28.0 & 14.6 & 7.9 & 13.4 & 29.4 & 46.0 
                 \\
                 ~ & NLX-GPT2$^{*}$ \cite{sammani2022nlx}
                 & CLIP-GPT2 & ~ & 54.4  & \underline{53.6}  & \underline{29.3} & \underline{15.5} & 8.4 & 13.6 & \underline{30.0} & \underline{53.8} 
                \\
                \cmidrule{2-12}
                 ~ & Baseline & CLIP-GPT2 & \underline{63.5} & \underline{58.9} & 53.2 & 29.2 & \underline{15.5} & 8.4 & 13.3 & 29.9 & 47.6 \\
                ~ & \textbf{SEVLM (Ours)} & CLIP-GPT2 & \textbf{65.6} & \textbf{62.1} &  \textbf{54.2} &  \textbf{30.3} & \textbf{16.4} & \textbf{9.2} & \textbf{13.9} & \textbf{30.4}  & \textbf{62.0} 
                \\
                    \midrule
                 \multirow{4}{*}{v2.0} &  SAT$^{*}$ \cite{achlioptas2021artemis}  & CNN-LSTM &  \multirow{2}{*}{\underline{43.3}} & \underline{38.8} & 48.7  & 25.3 & 13.2 & 7.3 & 12.8 & 27.2  & 57.0 
                \\
                ~ &  NLX-GPT2$^{*}$ \cite{sammani2022nlx} &  CLIP-GPT2 & ~ & 36.9 & 50.9 & 28.8 &  16.0 & \underline{9.2} & 13.6 & \underline{29.9} & 34.1 
                \\
                 \cmidrule{2-12}
                 ~ & Baseline & CLIP-GPT2 & 40.7 & 38.7 & \underline{51.2} & \underline{29.4} & \underline{16.4} & \underline{9.2} & \underline{13.7} & \underline{29.9} & \underline{61.2} \\
                ~& \textbf{SEVLM (Ours)} & CLIP-GPT2 & \textbf{44.2} & \textbf{42.6} & \textbf{51.8} & \textbf{30.4} & \textbf{17.2} & \textbf{10.1} & \textbf{13.9} & \textbf{30.4}  & \textbf{63.6} 
                \\
                \bottomrule
		\end{tabular}
   \end{threeparttable}
  \label{tab: Comparison on V1 and V2}
\end{table*}
\begin{table*}[t]
	\caption{Comparing SEVLM with LLaVA 7B after fine-tuning (denoted as LLaVA-FT). Our model is very competitive in most metrics, especially ACC and EA which are important indicators, except that LLaVA-FT has significantly higher diversity (lower Unique score). Moreover, our model is \textbf{37.5 times} smaller than LLaVA-FT. 
 }
	\centering
\scriptsize
   \begin{tabular}{c|l|c| c|c| ccccccc}
                    \toprule
                   Dataset & {Method} & {Backbone} & ACC$\uparrow$& EA$\uparrow$ &  B1$\uparrow$ &   B2$\uparrow$  & B3$\uparrow$ & B4$\uparrow$ & M$\uparrow$ & R$\uparrow$ & Unique$\uparrow$
                   \\
                    \midrule
                   \multirow{2}{*}{v1.0} & LLaVA-FT$^{*}$ \cite{liu2023llava} & CLIP-Vicuna & 
                   \underline{61.8} & \underline{60.0} & \textbf{55.5} & \textbf{30.7} & \textbf{16.4} & 9.0 & \textbf{14.3} & \textbf{30.7} & \textbf{84.3} \\ 
                ~ & \textbf{SEVLM (ours)} & {CLIP-GPT2} & \textbf{65.6} & \textbf{62.1} &  \underline{54.2} &  \underline{30.3} & \textbf{16.4} & \textbf{9.2} & \underline{13.9} & \underline{30.4}  & \underline{62.0} 
                \\ 
                 \midrule
                \multirow{2}{*}{v2.0} & LLaVA-FT$^{*}$ \cite{liu2023llava} & CLIP-Vicuna & 
                38.4 & \underline{39.3} & \textbf{53.2} & \textbf{30.6} & \underline{16.9} & \underline{9.5} & \textbf{14.3} & \textbf{30.7} & \textbf{84.5} \\      
                ~ & \textbf{SEVLM (ours)} & CLIP-GPT2 & 
                \textbf{44.2} & \textbf{42.6} & \underline{51.8} & \underline{30.4} & \textbf{17.2} & \textbf{10.1} & \underline{13.9} & \underline{30.4}  & \underline{63.6} \\
            \bottomrule
		\end{tabular}
   \label{tab: Comparison with LLaVA}
\end{table*}
\textbf{Comparison with the baseline.} {In \cref{tab: Comparison on V1 and V2}, we compare our method with the baseline on the two datasets and observe significant improvements. For example, ACC and EA of our method is +2.1\% and +3.2\% higher than the baseline on the ArtEmis v1.0 test set, respectively. The same metrics of our method is +3.5\% and +3.9\% higher on the ArtEmis v2.0 Combined test set, respectively. 

\textbf{Comparison with the state of the art.} In~\cref{tab: Comparison on V1 and V2}, we present the comparison with existing methods on the two affective datasets. Both two-stage methods \cite{achlioptas2021artemis,sammani2022nlx} consistently use ResNet-34 \cite{he2016deep} as the image emotion recognition model, and their captioning models refer distinct backbones listed in \cref{tab: Comparison on V1 and V2}. We clearly observe that our method significantly improves both emotion metrics (ACC and EA) and semantic metrics (\eg, B4 and R) on the two datasets. For example, on the ArtEmis v1.0 test set, our method bring remarkable improvements, \ie, +5.4\%,  +7.7\%, +0.8\%, and +0.4\% under ACC, EA, B4, and R, respectively. Our model also has the best performance on the Unique metric, with $+8.2\%$ and $+6.6\%$ improvement on both datasets, respectively, compared with the second best method. This is because as the model's ability in emotion identification improves, it can learn more subjective and personal interpretations rather than objective description paradigms.

\textbf{Comparison with LLaVA after fine-tuning.} In \cref{{tab: Comparison with LLaVA}}, we compare with LLaVA fine-tuned on the two datasets. We have the following observations. First, our model is very competitive compared with LLaVA-FT. ACC and EA of our method is consistently higher, while results are mixed but quite close for other metrics. Second, LLaVA-FT has much higher diversity in their language explanations than our method, which can be attributed to its much stronger pre-trained language models. Overall speaking, these results indicate the feasibility of our design as a competitor to much larger ones in emotion understanding.

\textbf{Computational efficiency comparisons.} \cref{tab: Computational Cost} summarizes the comparisons with the baseline and LLaVA-FT. Compared with the baseline, our model adds marginally more parameters and longer inference speed. Both can be trained and tested using a single 2080Ti. Compared with LLaVA-FT, our model is 37.5 times smaller and consumes much less computational resources and less floating point operations (FLOPs), while having a higher inference speed. Together with the emotion understanding performance comparisons above, we have achieved a much better accuracy-efficiency trade-off for small vision language models.
\begin{table*}[t]
	\caption{Computational efficiency comparison of different models.}
	\centering
 {
 \scriptsize
  \begin{tabular}{l|c c c c c c c}
             \toprule
           Model & \#Params  & Training GPU  & Training Time  & FLOPs$\downarrow$ & Inference GPU &  Inference Speed$\uparrow$
           \\
           \midrule
           Baseline & 168M & 1$\times$2080Ti   &7.5h, 15 epochs & 19.33G &1$\times$2080Ti & 8.71 (FPS) \\
           LLaVA-FT  & {7B} & {4$\times$4090} & 8h, 2 epoch
          & 4885.89G  & 1$\times$A40 & 3.89 (FPS) \\
            SEVLM & 186.45M & 1$\times$2080Ti  & 18h, 35 epochs & 19.33G & 1$\times$2080Ti &  8.62 (FPS)  \\
            \bottomrule
		\end{tabular}}
  \label{tab: Computational Cost}
\end{table*}

\label{sec: Further Analysis}
\begin{figure}[t]
	\centering
 \includegraphics[width=1\columnwidth]{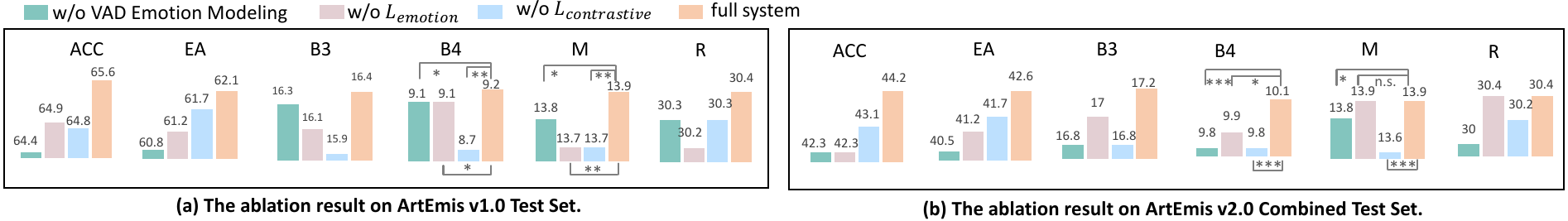}
  \caption{Ablation study of the three components on the ArtEmis v1.0 test set (a) and ArtEmis v2.0 combined test set(b). {We also perform statistical tests with p-value on B4 and M metrics, where the p-value is a statistic used to evaluate whether the difference in performance between two methods is significant.} `n.s.' means the differences is not statistically significant (\ie, p-value \textgreater 0.05). $\ast$ denotes statistically significant (\ie, 0.01 \textless p-value \textless 0.05). $\ast\ast$ and $\ast\ast\ast$ mean statistically very significant (\ie, 0.001 \textless p-value \textless 0.01) and statistically extremely significant (\ie, p-value \textless 0.001), respectively.}. 
	\label{tab: Ablation on V1 and V2}
\end{figure}
%\noindent
\begin{figure}[t]
  \centering
  \begin{minipage}[b]{0.48\textwidth}
    \includegraphics[width=\textwidth]{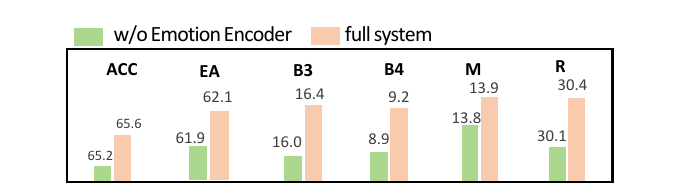}
    \caption{Ablation study of emotion encoder on ArtEmis v1.0 test set. `w/o Emotion Encoder' denotes that the VAD vectors extracted from the emotion dictionary are directly set as emotion features (in \cref{eq: emotion-enhanced f_X} ) without encoding them by the emotion encoder. 
    }
\label{fig:emotion_encoder}
  \end{minipage}
  \hfill
  \begin{minipage}[b]{0.48\textwidth}
\includegraphics[width=\textwidth]{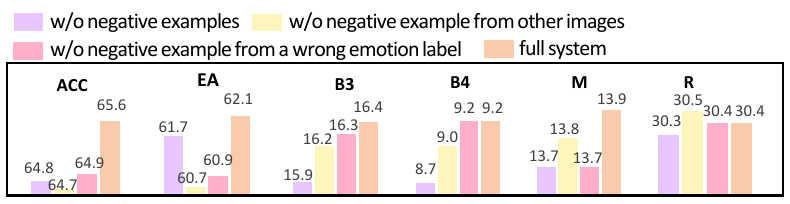}
    \caption{Negative samples ablation in contrastive loss on ArtEmis v1.0 test set. These two types of negative samples not only contribute to emotional alignment but also help achieve better performance in semantic metrics.}
    \label{fig:ablation_neg}
  \end{minipage}
\end{figure}

%\begin{figure}[t]
%	\centering
%	\includegraphics[width=0.99\columnwidth]{figs/Fig5_weightCLloss.pdf}
%	\caption{Impact of hyperparameter $\alpha$ defined in \cref{eq: all losses} on the ArtEmis v2.0 combined validation set. Left to right: metrics ACC, EA, B4, M, and R are used. We set $\alpha=2$ for both ArtEmis v2.0 combined and ArtEmis v1.0 validation set sets.}
%	\label{fig:weights}
%\end{figure}

\subsection{Further Analysis} 
\textbf{Main ablation studies.} Our system has three major improvements: VAD emotion features, VAD head, and contrastive head. We remove them one at time from the full system. Results are summarized in \cref{tab: Ablation on V1 and V2}. On the two test sets, we observe that removing any of the three components leads to performance drop in ACC and EA, two most important accuracy metrics, as well as semantic metrics such as BLEU and METEOR. Note these are confirmed by our statistical tests. These experiments indicate their effectiveness.

\textbf{Detailed ablations and variants of the VAD emotion feature and contrastive head.} We present results in \cref{fig:emotion_encoder} and \cref{fig:ablation_neg}, for the two components, respectively. For the \underline{VAD emotion feature}, if we remove the emotion encoder (see \cref{sec: VAD Embeddings} for its description), all metrics on the ArtEmis V1.0 test set become worse, as shown in \cref{fig:emotion_encoder}. Moreover, the \underline{contrastive head} use two types of negative samples as shown in \cref{fig4:detailed network}; if we remove either of two, we also observe performance drop in \cref{fig:ablation_neg}. These ablation and variant studies further verify the reasonableness of our system. More experimental results, \eg, the impact of hyperparameter $\alpha$ and more analysis with existing methods are included in the supplementary material.

\begin{figure*}[t]
	\centering
	\includegraphics[width=0.99\linewidth]{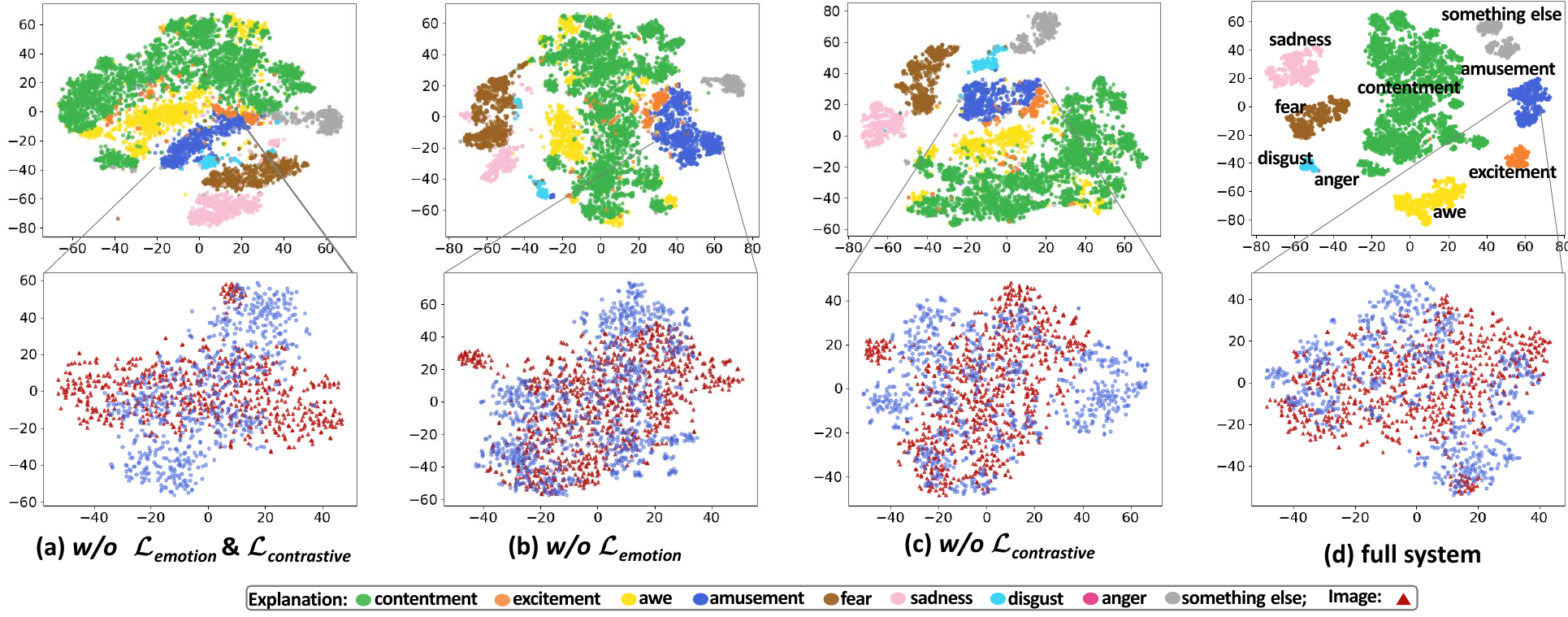}
	\caption{T-SNE visualization of feature distributions on ArtEmis v1.0 test set. (\textbf{Top:}) features of language explanations. Different colors denote their corresponding emotion classes. Compared with the three variants (a), (b), and (c), the full model better separates explanations of different classes. (\textbf{Bottom:}) features of explanations (blue) and corresponding images (red), exemplified by emotion `amusement'. Our method demonstrates better alignment of the distributions of the two features.}  
	\label{fig:tsne}
\end{figure*}
\begin{figure}[t]
	\centering
	\includegraphics[width=0.99\columnwidth]{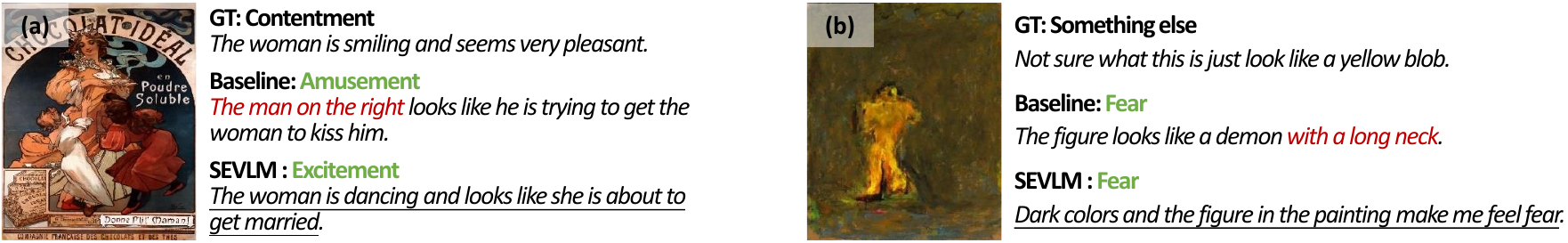}
	\caption{Failure cases: (a) on ArtEmis v1.0 test set and (b) on ArtEmis v2.0 Combined test set. We find our emotion category and explanation predictions to some extent align with the image, but the predicted emotion class is different from the ground truth label. This may be attributed to the subjectivity and ambiguity of emotion classification.} 
	\label{fig:failure cases}
\end{figure}
\subsection{Qualitative Results}
\textbf{Visualization.} In~\cref{fig1:comparison}, we give visualization examples of different models on two datasets. We found that existing methods have insufficient performance in emotional understanding and can be divided into two types. In one case, when the emotion classification is correct, semantic errors will occur in the interpretation.  For example, in \cref{fig1:comparison} (a), SAT and NLX-GPT2 describe the wrong action `dancing' and wrong object `horse' respectively, while our method generates the correct explanation `the man is fighting with the man.' The other is when the emotion prediction is wrong, the interpretation will not only have semantic errors but also a mismatch between the interpretation and the emotion category. For example, in \cref{fig1:comparison} (c), the category `Contentment' predicted by NLX-GPT2 is inconsistent with explanation, \ie, `the woman looks like she is about to cry.' These visualization comparison demonstrates the superiority of our approach for emotion understanding of artistic images. See more examples in our supplementary.

\textbf{Feature distribution visualisation.} We use t-SNE to visualize the feature distributions of language explanations (denoted with category labels) and images in \cref{fig:tsne}. Compared with models that lack certain components, the full system clearly has explanations that are better separated according to different emotion classes, \eg, `contentment', `excitement', and `awe'. We further visualize the explanation and image features in the `amusement' category and observe that the two types of features align better under the full system. These visualizations demonstrate the effectiveness of our system in aligning the art image with emotion classes and explanations.

\textbf{Failure cases.} The predicted emotion classes and explanations in~\cref{fig:failure cases} are judged as failure cases by ground-truth labels. However, they look reasonable to some extent. In (a), the emotion prediction `Amusement' is close to the ground truth `Contentment', and the explanations are reasonable. In (b), our model predicts the emotion as `Fear', and the explanation also aligns with this prediction and image. So we speculate that many failure cases are due to the nature of affective computing, where there is no objective standard.

\begin{table*}[t]	
\caption{Comparison results of GPT4(V) and our SEVLM on random 100 samples of ArtEmis v1.0 test set.} 
	\centering
 \scriptsize
		\begin{tabular}{ c | c | c| c c c c c c | c  }
                    \toprule
                     Method & ACC & Emo-Align & B1 & B2 & B3 & B4 & M & R  & Unique \\
                    \midrule
                  GPT4(V) & \underline{52.6} & \underline{43.8} & \underline{31.5} & \underline{10.2} & \underline{3.3} & \underline{1.3} & \underline{8.4} & \underline{20.4}  & \textbf{100.0} \\
                SEVLM & \textbf{63.2} & \textbf{56.3} &   \textbf{54.8} & \textbf{30.7} & \textbf{17.9} & \textbf{11.4} & \textbf{14.2} & \textbf{29.8} & \underline{96.0}  \\
                 \bottomrule
		\end{tabular}
    \label{tab: Comparison results of GPT4}
\end{table*}

\begin{figure*}[t]
	\centering
	\includegraphics[width=0.99\linewidth]{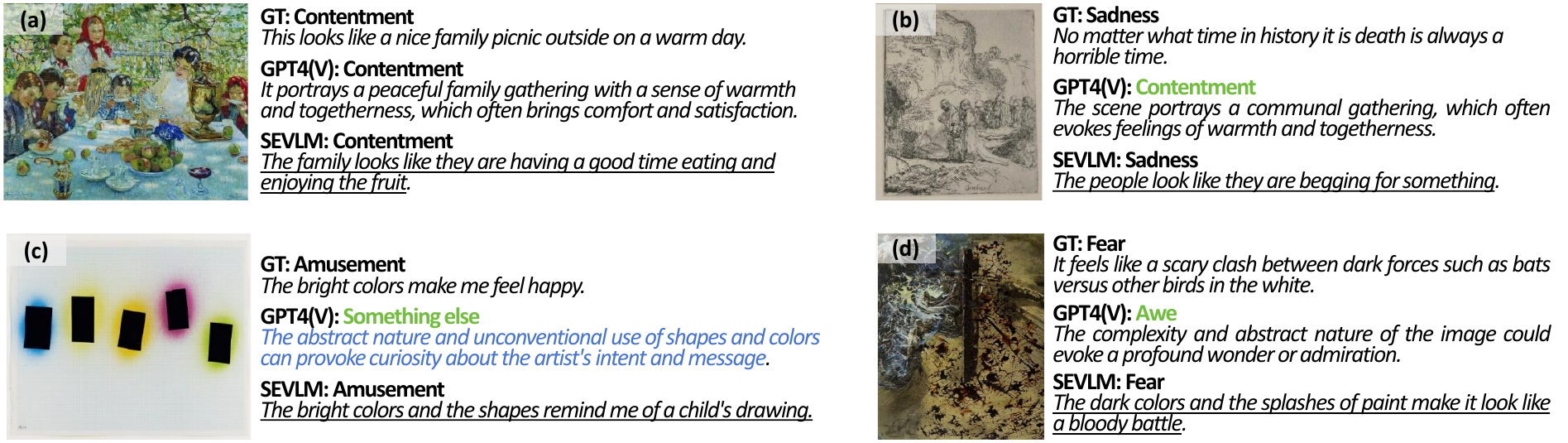}
	\caption{Examples of GPT4(V) and our SEVLM on ArtEmis v1.0 test set. \textcolor[RGB]{117,189,66}{\textbf{Green fonts}} indicate incorrect emotion results; \textcolor[RGB]{72,116,203}{\textbf{blue texts}} denote that the emotion of the explanations does not correspond to the predicted category.}
	\label{fig:GPT4_examples} 
\end{figure*}
\subsection{Discussions} 
To \textbf {compare with GPT4(V)}, we randomly sampled 100 examples from the test set of ArtEmis v1.0 to evaluate the emotion understanding performance of GPT4(V) \footnote{The ArtEmis v1.0 test set contains nearly 7K data and GPT4(V) requires payment. Considering the cost and time, we randomly selected 100 experiment samples available at \href{https://github.com/BetterZH/SEVLM-code}{here}.}. As shown in~\cref{tab: Comparison results of GPT4}, our model significantly outperforms GPT4(V) in almost all emotion metrics and semantics metrics except the Unique metric. In addition, we present qualitative results in~\cref{fig:GPT4_examples}. We observe that GPT4(V) emphasizes objective content, lacking subjective perceptual experiences. For example, in \cref{fig:GPT4_examples} (a) and (b) which are both related to `gathering', GPT4(V) fails to capture the emotional differences between these two artworks, while our model generates correct emotions from an individual perspective and provides reasonable explanations. Moreover, GPT4(V) faces challenges in emotional perception on abstract artworks. In comparison, our model can accurately comprehend the emotions evoked by abstract paintings and imaginatively associate abstract elements, such as `the dark colors and the splashes of paint make it look like a bloody battle' in~\cref{fig:GPT4_examples} (d). 

\textbf{Can LLaVA-FT be improved by the proposed techniques?} We tried implementing the proposed methods to LLaVA using LoRA \cite{hu2021lora} for fine-tuning, but we did not observe noticeable performance improvement. We speculate that the deep structures in LLaVA compromises proper gradient propagation and parameter update under our techniques.{An expected approach involves implementing suitable model compression techniques for large models to simplify the network, making it more conducive to deploying our technology. However, the exploration of model compression for large models is presently a focal point of research, and there is no widely acknowledged universal algorithm. This presents itself as a subject that could be pursued as an independent research direction, going beyond the scope of this paper.}

\textbf{How important is the Unique metric?} This metric measures diversity, which is very different from other metrics in emotion explanation and classification which focus on alignment with the ground truth, or accuracy. The Unique metric may be prone to the hallucination problem inherent in strong language decoders such as LLaVA. So what might happen is a very high diversity but compromised emotion analysis accuracy. Our opinion would be prioritizing the accuracy metrics and view diversity as a secondary goal.

\section{Conclusion}
In this paper, we propose a small generative model for emotion recognition and emotion-grounded explanation for artworks, where GPT2 is used as the backbone decoder. This model consistently outperforms the state-of-the-art small models in emotion understanding and is competitive with large models such as fine-tuned LLaVA and GPT4(V) while maintaining computational efficiency. The strong performance is achieved by 1) designing a contrastive head aligning image, prompt, and explanation features to reduce mismatches during inference, and 2) integrating the VAD modeling method into the input text embedding and the loss function to promote subjectivity in language explanations. In future work, we will explore various other domains where small models can be equally or more effective with large models and how existing human expert knowledge such as VAD modeling can be effectively integrated in large models.

\section*{Acknowledgements}
This work is supported by the National Natural Science Foundation of China (62272144, 72188101, 62020106007, and U20A20183), the Major Project of Anhui Province (202203a05020011), and the Fundamental Research Funds for the Central Universities (JZ2024HGTG0309).

% ---- Bibliography ----
%
% BibTeX users should specify bibliography style 'splncs04'.
% References will then be sorted and formatted in the correct style.
%
\bibliographystyle{splncs04}
\bibliography{main}
\end{document}